\documentclass[letterpaper, 10 pt, conference]{class/ieeeconf}  

\IEEEoverridecommandlockouts                              
\usepackage{amsmath}
\usepackage{amssymb}
\usepackage{latexsym}
\usepackage{url}
\usepackage{cite}
\usepackage{relsize}
\usepackage{multirow}
\usepackage{afterpage}
\usepackage{color}
\usepackage{ifthen}
\usepackage{graphicx}
\usepackage{algpseudocode,algorithm,algorithmicx}
\usepackage{subfigure}
\usepackage[keeplastbox]{flushend}
\usepackage{epstopdf}
\usepackage{gensymb}
\usepackage[bookmarks=false, linkcolor=blue, urlcolor=blue, citecolor=blue]{hyperref} 
\usepackage{booktabs}
\usepackage{makecell}
\usepackage{hhline}
\usepackage{lipsum}
\usepackage{mathtools}

\usepackage{subfloat}
\usepackage{booktabs}
\usepackage{varwidth}
\usepackage{capt-of}

\hypersetup{pdfstartview={FitH}}

\title{\LARGE \bf Real-Time 6DOF Pose Relocalization for Event Cameras with \\Stacked Spatial LSTM Networks}

\author{Anh Nguyen$^{1}$, Thanh-Toan Do$^{2}$, Darwin G. Caldwell$^{1}$, and Nikos G. Tsagarakis$^{1}$%
        \thanks{$^{1}$Anh Nguyen, Darwin G. Caldwell, Nikos G. Tsagarakis are with the Department of Advanced Robotics, IIT, Italy. {\tt \{Anh.Nguyen, Darwin.Caldwell, Nikos.Tsagarakis\}@iit.it}}
        \thanks{$^{2}$Thanh-Toan Do is with the Australian Centre for Robotic Vision, University of Adelaide. {\tt thanh-toan.do@adelaide.edu.au}}
}

\begin{document}

\newtheorem{problem}{Problem}
\newtheorem{lemma}{Lemma}
\newtheorem{theorem}[lemma]{Theorem}
\newtheorem{claim}{Claim}
\newtheorem{corollary}[lemma]{Corollary}
\newtheorem{definition}[lemma]{Definition}
\newtheorem{proposition}[lemma]{Proposition}
\newtheorem{remark}[lemma]{Remark}
\newenvironment{LabeledProof}[1]{\noindent{\it Proof of #1: }}{\qed}

\def\beq#1\eeq{\begin{equation}#1\end{equation}}
\def\bea#1\eea{\begin{align}#1\end{align}}
\def\beg#1\eeg{\begin{gather}#1\end{gather}}
\def\beqs#1\eeqs{\begin{equation*}#1\end{equation*}}
\def\beas#1\eeas{\begin{align*}#1\end{align*}}
\def\begs#1\eegs{\begin{gather*}#1\end{gather*}}

\newcommand{\poly}{\mathrm{poly}}
\newcommand{\eps}{\epsilon}
\newcommand{\e}{\epsilon}
\newcommand{\polylog}{\mathrm{polylog}}
\newcommand{\rob}[1]{\left( #1 \right)} 
\newcommand{\sqb}[1]{\left[ #1 \right]} 
\newcommand{\cub}[1]{\left\{ #1 \right\} } 
\newcommand{\rb}[1]{\left( #1 \right)} 
\newcommand{\abs}[1]{\left| #1 \right|} 
\newcommand{\zo}{\{0, 1\}}
\newcommand{\zonzo}{\zo^n \to \zo}
\newcommand{\zokzo}{\zo^k \to \zo}
\newcommand{\zot}{\{0,1,2\}}
\newcommand{\en}[1]{\marginpar{\textbf{#1}}}
\newcommand{\efn}[1]{\footnote{\textbf{#1}}}
\newcommand{\vecbm}[1]{\boldmath{#1}} 
\newcommand{\uvec}[1]{\hat{\vec{#1}}}
\newcommand{\thv}{\vecbm{\theta}}
\newcommand{\junk}[1]{}
\newcommand{\var}{\mathop{\mathrm{var}}}
\newcommand{\rank}{\mathop{\mathrm{rank}}}
\newcommand{\diag}{\mathop{\mathrm{diag}}}
\newcommand{\tr}{\mathop{\mathrm{tr}}}
\newcommand{\acos}{\mathop{\mathrm{acos}}}
\newcommand{\atantwo}{\mathop{\mathrm{atan2}}}
\newcommand{\SVD}{\mathop{\mathrm{SVD}}}
\newcommand{\quadf}{\mathop{\mathrm{q}}}
\newcommand{\linterp}{\mathop{\mathrm{l}}}
\newcommand{\sgn}{\mathop{\mathrm{sign}}}
\newcommand{\sym}{\mathop{\mathrm{sym}}}
\newcommand{\avg}{\mathop{\mathrm{avg}}}
\newcommand{\mean}{\mathop{\mathrm{mean}}}
\newcommand{\erf}{\mathop{\mathrm{erf}}}
\newcommand{\grad}{\nabla}
\newcommand{\R}{\mathbb{R}}
\newcommand{\defeq}{\triangleq}
\newcommand{\dims}[2]{[#1\!\times\!#2]}
\newcommand{\sdims}[2]{\mathsmaller{#1\!\times\!#2}}
\newcommand{\udims}[3]{#1}
\newcommand{\udimst}[4]{#1}
\newcommand{\com}[1]{\rhd\text{\emph{#1}}}
\newcommand{\ind}{\hspace{1em}}
\newcommand{\argmin}[1]{\underset{#1}{\operatorname{argmin}}}
\newcommand{\floor}[1]{\left\lfloor{#1}\right\rfloor}
\newcommand{\step}[1]{\vspace{0.5em}\noindent{#1}}
\newcommand{\quat}[1]{\ensuremath{\mathring{\mathbf{#1}}}}
\newcommand{\norm}[1]{\left\lVert#1\right\rVert}
\newcommand{\ignore}[1]{}
\newcommand{\specialcell}[2][c]{\begin{tabular}[#1]{@{}c@{}}#2\end{tabular}}
\newcommand*\Let[2]{\State #1 $\gets$ #2}
\newcommand{\algorithmicbreak}{\textbf{break}}
\newcommand{\Break}{\State \algorithmicbreak}
\newcommand{\ra}[1]{\renewcommand{\arraystretch}{#1}}

\renewcommand{\vec}[1]{\mathbf{#1}} 

\algdef{S}[FOR]{ForEach}[1]{\algorithmicforeach\ #1\ \algorithmicdo}
\algnewcommand\algorithmicforeach{\textbf{for each}}
\algrenewcommand\algorithmicrequire{\textbf{Require:}}
\algrenewcommand\algorithmicensure{\textbf{Ensure:}}
\algnewcommand\algorithmicinput{\textbf{Input:}}
\algnewcommand\INPUT{\item[\algorithmicinput]}
\algnewcommand\algorithmicoutput{\textbf{Output:}}
\algnewcommand\OUTPUT{\item[\algorithmicoutput]}

\maketitle
\thispagestyle{empty}
\pagestyle{empty}

\begin{abstract}
We present a new method to relocalize the 6DOF pose of an event camera solely based on the event stream. Our method first creates the event image from a list of events that occurs in a very short time interval, then a Stacked Spatial LSTM Network (SP-LSTM) is used to learn the camera pose. Our SP-LSTM is composed of a CNN to learn deep features from the event images and a stack of LSTM to learn spatial dependencies in the image feature space. We show that the spatial dependency plays an important role in the relocalization task and the SP-LSTM can effectively learn this information. The experimental results on a publicly available dataset show that our approach generalizes well and outperforms recent methods by a substantial margin. Overall, our proposed method reduces by approx. $6$ times the position error and $3$ times the orientation error compared to the current state of the art. The source code and trained models will be released.

\end{abstract}

\section{INTRODUCTION} \label{Sec:Intro}
Inspired by human vision, the event-based cameras asynchronously capture an event whenever there is a brightness change in a scene\cite{Brandli2014}. An event is simply composed of a pixel coordinate, its binary polarity value, and the timestamp when the event occurs. This differs from the frame-based cameras where an entire image is acquired at a fixed time interval. Based on its novel design concept, the event-based camera can rapidly stream events (i.e., at microsecond speeds). This is superior to frame-based cameras which usually sample images at millisecond rates~\cite{Censi2014}. This novel ability makes the event cameras more suitable for the high-speed robotic applications that require low latency and high dynamic range from the visual data.

Although the event camera creates a paradigm shift in solving real-time visual problems, its data come extremely quickly without the intensity information usually found in an image. Each event also carries very little information (i.e., the pixel coordinate, the polarity value and the timestamp) when it occurs. Therefore, it is not trivial to apply standard computer vision techniques to event data. Recently, the event camera is gradually becoming more popular in the computer vision and robotics community. Many problems such as camera calibration and visualization~\cite{Mueggler2016_dataset}, 3D  reconstruction~\cite{Kim2016}, simultaneous localization and mapping (SLAM)~\cite{Weikersdorfer2014}, and pose tracking~\cite{Mueggler2014_posetracking} have been actively investigated.


\begin{figure}[!t] 
    \centering

\includegraphics[width=0.85\linewidth, height=0.60\linewidth]{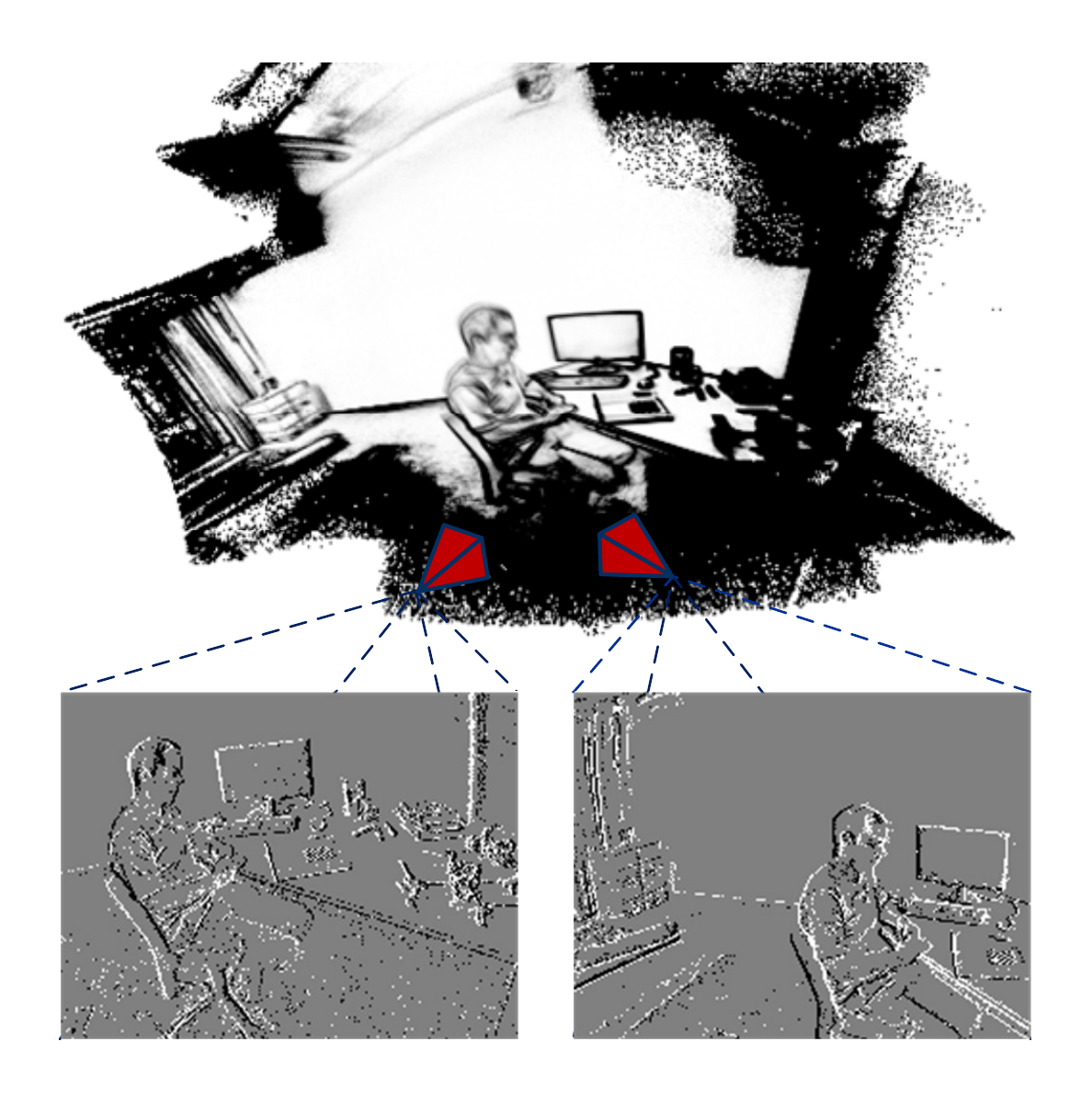}   		
    \vspace{1.0ex}
    \caption{Pose relocalization for event cameras. We propose to create event images from lists of events and relocalize the 6DOF camera poses from these images using a deep neural network.}
    \label{Fig:intro} 
\end{figure}

Our goal in this work is to develop a new method, which relocalizes the 6 Degrees of Freedom (6DOF) pose of the event camera using a deep learning approach. The problem of effectively and accurately interpreting the pose of the camera plays an important role in many robotic applications such as navigation and manipulation. However, in practice it is challenging to estimate the pose of the event camera since it can capture a lot of events in a short time interval, yet each event does not have enough information to perform the estimation. We propose to form a list of events into an event image and regress the camera pose from this image with a deep neural network. The proposed approach can accurately recover the camera pose directly from the input events, without the need for additional information such as the 3D map of the scene or inertial measurement data.

In computer vision, Kendall et al.~\cite{Alex2015} introduced a first deep learning framework to retrieve the 6DOF camera pose from a single image. The authors in~\cite{Alex2015} showed that compared to the traditional keypoint approaches, using CNN to learn deep features resulted in a system that is more robust in challenging scenarios such as noisy or uncleared images. Recently, the work in~\cite{Kendall2017} introduced a method that used a geometry loss function to learn the spatial dependencies. In this paper, we employ the same concept, using CNN to learn deep features, however unlike~\cite{Kendall2017} that builds a geometry loss function based on the 3D points in the scene, we use an SP-LSTM network to encode the geometry information. Our approach is fairly simple but shows significant improvement over the state-of-the-art methods.

The rest of the paper is organized as follows. We review related work in Section~\ref{Sec:rw}, followed by a description of the event data and event images in Section~\ref{Sec:event_data}. The SP-LSTM network is introduced in Section~\ref{Sec:splstm}. In Section~\ref{Sec:exp}, we present the extensive experimental results. Finally, we conclude the paper and discuss the future work in Section~\ref{Sec:con}.


\junk{

due to the increasing or decresing of the local brightness

Moreover, we also analyze the effect of the number events that contribute to the pose estimation result.

15-30 Hz

Code for our models and metrics, as well as dataset fea- tures and predictions,1 will be released upon acceptance.

First, a single event does not contain enough information to estimate the sensor pose given by the six degrees of freedom (DOF) of a calibrated camera. Additionally, it is not appropriate to simply consider several events for determining the pose  using standard computer-vision techniques, such as PnP [4], because the events typically all have different timestamps, and so the resulting pose will not correspond to any particular time. Second, an event camera can easily transmit up to several million events per second, and, therefore, it can become intractable to estimate the pose of the event camera at the discrete times of all events due to the rapidly growing size of the state vector needed to represent all such poses

Event cameras offer a breakthrough new paradigm for real-time vision, with po- tential in robotics, wearable devices and autonomous vehicles, but it has proven very challenging to use them in most standard computer vision problems. In- spired by the superior properties of human vision [2], an event camera records not image frames but an asynchronous sequence of per-pixel intensity changes, each with a precise timestamp. While this data stream efficiently encodes image dynamics with extremely high dynamic range and temporal contrast, the lack of synchronous intensity information means that it is not possible to apply much of the standard computer vision toolbox of techniques. In particular, themulti-view correspondence information which is essential to estimate motion and structure is difficult to obtain because each event by itself carries little information and no signature suitable for reliable matching

The event camera or silicon retina is gradually becoming more widely known by researchers in computer vision, robotics and related fields, in particular since the release as a commercial device for researchers of the Dynamic Vision Sensor (DVS) [14] shown in Figure 1 (c). The pixels of this device asynchronously re- port log intensity changes of a preset threshold size as a stream of asynchronous events, each with pixel location, polarity, and microsecond-precise timestamp. Figure 1 visualises some of the main properties of the event stream; in particular the almost continuous response to very rapid motion and the way that the output data-rate depends on scene motion, though in practice almost always dramatically lower than that of standard video. These properties offer the potential to overcome the limitations of real-world computer vision applications, relying on conventional imaging sensors, such as high latency, low dynamic range, and high power consumption.

One possible alternative is to use a Dynamic Vision Sensor (DVS) [4]. This is the first commercially available product belonging to a new class of “neuromorphic” sensors [5, 6]. In contrast to a normal CMOS camera, the DVS output is a sequence of asynchronous events rather than regular frames. Each pixel produces an event when the perceived luminance increases and decreases under a certain threshold. This computation is done using an analog circuit, whose biases can be tuned to change the sensitivity of the pixels and other dynamic properties. The events are then timestamped and made available to the application using a digital circuit

In this study, The context is important for object recognition. Our contributions are threefold:

\begin{itemize}
\item A deep learning based method that can be trained end-to-end to directly interpret input videos to robot commands.

\item A dataset that is sufficient for deep learning methods.

\item An robotic framework that use the proposed method to execute different manipulation tasks.

\end{itemize} 

\begin{figure*}[!t] 
    \centering
	\includegraphics[width=0.99\linewidth, height=0.2\linewidth]{example-image-a} 
    \vspace{2.0ex}
    \caption{TBD.}
    \label{Fig:overview} 
\end{figure*}

research about event camera has received more and more interests in the computer vision and robotics community~\cite{•}~\cite{•}~\cite{•}.

Standard frame-based CMOS cameras operate at fixed frame rates, sending entire images at constant time intervals that are selected based on the considered application [1]. Contrary to standard cameras, where pixels are acquired at regular time intervals (e.g., global shutter or rolling shutter), event cameras, such as Dynamic Vision Sensors (DVS [2] or DAVIS [3]), have asynchronous, independent pixels: each pixel of an event camera immediately triggers an event whenever it detects a brightness change in the scene (Fig. 1). The temporal resolution of such events is in the order of micro-seconds. It is only the sign of these changes that is transmitted, which is binary (increase or decrease of local brightness). Because the output it produces—an event stream—is fundamentally differ- ent from video streams of standard cameras, new algorithms are required to deal with these data.

}

\section{Related Work} \label{Sec:rw}
The event camera is particularly suitable for real-time motion analysis or high-speed robotic applications since it has low latency~\cite{Mueggler2016_dataset}. Early work on event cameras used this property to track an object to provide fast visual feedback to control a simple robotic system~\cite{Conradt2009}. The authors in~\cite{Mueggler2014_posetracking} set up an onboard perception system with an event camera for 6DOF pose tracking of a quadrotor. Using the event camera, the quadrotor's poses can be estimated with respect to a known pattern during high-speed maneuvers. Recently, a 3D SLAM system was introduced in~\cite{Weikersdorfer2014} by fusing frame-based RGB-D sensor data with event data to produce a sparse stream of 3D points. This sparse stream is a compact representation of the input events, hence it uses less computational resource and enables fast tracking.

In~\cite{Kim2014}, the authors presented a method to estimate the rotational motion of the event camera using two probabilistic filters. Recently, Kim et al.~\cite{Kim2016} extended this system with three filters that simultaneously estimate the 6DOF pose of the event camera, the depth, and the brightness of the scene. The work in~\cite{Gallego2017} introduced a method to directly estimate the angular velocity of the event camera based on a contrast maximization design without requiring optical flow or image intensity estimation. Reinbacher et al.~\cite{Reinbacher2017} introduced a method to track an event camera based on a panoramic setting that only relies on the geometric properties of the event stream. More recently, the authors in~\cite{zihao2017event}~\cite{Rebecq2017} proposed to fused events with IMU data to accurately track the 6DOF camera pose.

In computer vision, 6DOF camera pose relocalization is a well-known problem. Recent research trends investigate the capability of deep learning for this problem~\cite{Alex2015}~\cite{Kendall2017}~\cite{Walch2016}. Kendall et al.~\cite{Alex2015} introduced a first deep learning framework to regress the 6DOF camera pose from a single input image. The work of~\cite{Alex2016} used Bayesian uncertainty to correct the camera pose. Recently, the authors in~\cite{Kendall2017} introduced a geometry loss function based on 3D points from a scene, to let the network encode the geometry information during the training phase. Walch et al.~\cite{Walch2016} used a CNN and four parallel LSTM together to learn the spatial relationship in the image feature space. The main advantage of the deep learning approach is that the deep network can effectively encode the features from the input images, without relying on the hand-designed features.

This paper follows the recent trend in computer vision by using a deep network to estimate the pose of the event camera. We first create an event image from a list of events. A deep network composed of a CNN and an SP-LSTM is then trained end-to-end to regress the 6DOF camera pose. Unlike~\cite{Kendall2017} that used only CNN with a geometry loss function that required the 3D points from the scene, or{~\cite{Walch2016} that used four parallel LSTM to encode the geometry information, we propose to use Stacked Spatial LSTM to learn spatial dependencies from event images. To the best of our knowledge, this is the first deep learning approach that successfully relocalizes the pose of the event camera.

\junk{
high-speed measurement and 

depth-augmented
event-based

\textbf{TBD: 6DOF pse estimation for event camera:} Although 6DOF pose estimation is a well-known problem, most of proposed methods are for the frame-based cameras~\cite{•}~\cite{•}. There are few works on 3D orientation estimation with event cameras. This may be due to the following facts: research is dominated by standard (frame-based) cameras, event cameras have been commercially available only recently [1] and they are still expensive sensors since they are at an early stage of development
Robot localization in 6-DOF with respect to a map of BW lines was demonstrated using a DVS, without additional sensing, during high-speed maneuvers of a quadrotor [19], where rotational speeds of up to 1,200 were measured. In natural scenes, [20] presented a probabilistic filter to track high-speed 6-DOF motions with respect to a map containing both depth and brightness information.

We present an algorithm to estimate the rotational
motion of an event camera. In contrast to traditional cameras, which produce images at a fixed rate, event cameras have independent pixels that respond asynchronously to brightness changes, with microsecond resolution. Our method leverages the type of information conveyed by these novel sensors (i.e., edges) to directly estimate the angular velocity of the camera, without requiring optical flow or image intensity estimation. The core of the method is a contrast maximization design. The method performs favorably against ground truth data and gyroscopic

\textbf{TBD: General event camera problem.} Early published work using event cameras focused on tracking moving objects from a fixed point of view, successfully showing the superior high speed mea- surement and low latency properties [8, 6]. In several works, conventional vision sensors have been attached to the event camera to simplify the ego-motion estimation problem. For example, [16] proposed an event- based probabilistic framework to update the relative pose of a DVS with respect to the last frame of an attached standard camera. The 3-D SLAM system in [17] relied on a frame- based RGB-D camera attached to the DVS to provide depth estimation, and thus build a voxel grid map that was used for pose tracking. The system in [18] used the intensity images from the DAVIS camera to detect features that were tracked using the events and were then fed into a 3-D visual odometry pipeline.

 However, unlike
the work in [6] that uses the CNN on the whole input image,
we employ an object detector to narrow down the region of
interest. In [5], Srikantha and Gall made the assumption that
the human pose is known and can be integrated into the
detection process. In this paper, we show that the detection
results can be improved by integrating an object detector
and refining the result with CRF. Furthermore, we introduce
a grasping method based on the concept of principle component
analysis and run extensive grasping experiments on
the full-size humanoid robot WALK-MAN

\textbf{TBD: Our method}All previous methods operate in an event-by-event basis producing estimates of the event camera pose in a discrete, filter-like manner. This paper leverages a continuous-time representation of the trajectory of the event camera to couple the estimated poses in a tractable batch optimization that also allows to fuse event and inertial data.

}

\section{Event Data} \label{Sec:event_data}
\subsection{Event Camera}
Instead of capturing an entire image at a fixed time interval as in standard frame-based cameras, the event cameras only capture a single event at a timestamp based on the brightness changes at a local pixel. In particular, an event $e$ is a tuple $e =  < e_t,({e_x},{e_y}),{e_\rho } >$ where $e_t$ is the timestamp of the event, $(e_x, e_y)$ is the pixel coordinate and $e_\rho  =  \pm 1$ is the polarity that denotes the brightness change at the current pixel. The events are transmitted asynchronously with their timestamps using a sophisticated digital circuitry. Recent event cameras such as DAVIS 240C~\cite{Brandli2014} also provide IMU data and global-shutter images. In this work, we only use the event stream as the input for our deep network.

\begin{figure}[ht] 
    \centering

	\includegraphics[width=0.99\linewidth, height=0.72\linewidth]{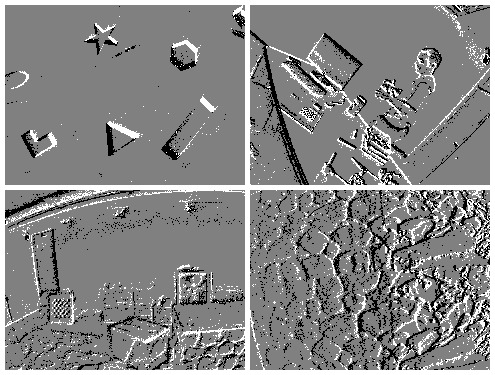}   		
    \vspace{0ex}
    \caption{Examples of event images. Since the events mainly occur around the edge of the objects, the event images are clearer on simple scenes (top row), while more disorder on cluttered scenes (bottom row).}
    \label{Fig:event_image} 
\end{figure}

\subsection{From Events to Event Images}

\begin{figure*}[!t] 
    \centering
	\includegraphics[width=0.99\linewidth, height=0.17\linewidth]{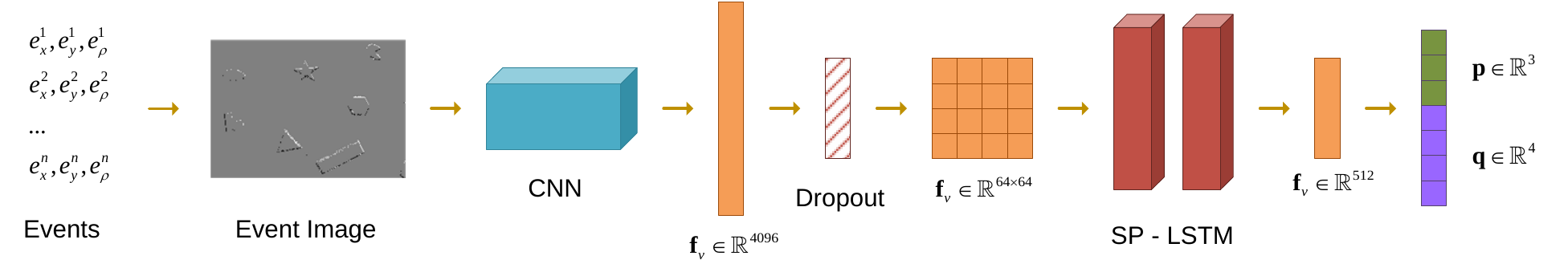}
    \hspace{2.0ex}
    \caption{An overview of our 6DOF pose relocalization method for event cameras. We first create an event image from a list of events, then a CNN is used to learn deep features from this image. The image feature vector is reshaped and fed to a SP-LSTM network with $256$ hidden units. Finally, the output of SP-LSTM is fed to a fully connected layer with $512$ neurons, following by another fully connected layer with $7$ neurons to regress the pose vector.}
    \label{Fig:overview} 
\end{figure*}

Since a single event only contains a binary polarity value of a pixel and its timestamp, it does not carry enough information to estimate the 6DOF pose of the camera. In order to make the pose relocalization problem using only the event data becomes feasible, similar to~\cite{Gallego2017} we assume that $n$ events in a very short time interval will have the same camera pose. This assumption is based on the fact that the event camera can capture many events in a short period, while in that very short time interval, the poses of the camera can be considered as unchanging significantly. From a list of $n$ events, we reconstruct an event image $I \in \mathbb{R}^{h \times w}$ (where $h$ and $w$ are the height and width resolution of the event camera) based on the value of the polarity $e_\rho$ as follows:

 \begin{equation}
    I(e_x,e_y)=
    \begin{cases}
      0, & \text{if}\ e_\rho=-1 \\
      1, & \text{if}\ e_\rho=1  \\
      0.5, & \text{otherwise}
    \end{cases}
  \end{equation}
  
This conversion allows us to transform a list of events to an image and apply traditional computer vision techniques to event data. Since the events mainly occur around the edge of the scene, the event images are clearer on simple scenes, while more disorder on cluttered scenes. Fig.~\ref{Fig:event_image} shows some examples of event images. In practice, the parameter $n$ plays an important role since it affects the quality of the event images, which are used to train and infer the camera pose. We analyze the effect of this parameter to the pose relocalization results in Section~\ref{Sec:exp}-D.

\junk{
A dropout layer is used after the CNN network to avoid overfitting.

In practice, we use the state-of-the-art deep CNN to extract these feature vectors from the input images.

$e_i = <t_i, (x_i, y_i), \rho_i>$
It also has higher high dynamic range in comparison with the frame-based cameras ($130$ dB vs. $60$ dB).

Since the network need to learn the translation and rotation of the pose vector at the same time, while these two components are measured in different units, it is not trivial to design an effective loss function. To train the network, we used the following objective loss function:

\begin{equation}
\mathcal{L}(I) = {\left\| {\mathbf {\hat x} - \mathbf{x}} \right\|_2} + \beta {\left\| {\mathbf{\hat q} - \frac{\mathbf{q}}{{\left\| \mathbf{q} \right\|}}} \right\|_2}
\end{equation}

where $\beta$ is a parameter to keep the loss from the position and orientation to be approximately equal. For simplicity, the orientation loss is calculated on Euclidean space instead of the unit sphere in quaternion space since in practice the distinction between spherical distance and Euclidean distance is insignificant~\cite{•}. In~\cite{•}, the authors introduced a new loss function based on the reprojection error. However, this approach needed a list of 3D points to measure the projection error with the estimated pose $\mathbf{\hat p}$, which is not available in the groundtruth of our dataset.

(three dimensions represent position $\mathbf{x}$ and four dimensions represent orientation $\mathbf{q}$),

In practice, the LSTM network is straightforward to train end-to-end and can handle inputs with different lengths using the padding techniques.

To demonstrate the generality of our approach, we instantiate Mask R-CNN with multiple architectures. For

This allows each layer in the mask branch to maintain the explicit m × m object spatial lay- out without collapsing it into a vector representation that lacks spatial dimensions. Unlike

Even though Long Short-Term Memory (LTSM) units
have been typically applied to temporal sequences, recent works [6, 32, 53, 54] have used the memory capabilities of LSTMs in image space. In our case, we treat the output 2048 feature vector as our sequence. We propose to insert four LSTM units after the FC, which have the function of reducing the dimensionality of the feature vector in a struc- tured way. The memory units filter out the most useful fea- ture correlations for the task of pose estimation

Stack of LSTM, on other hands, is simply a stack of several LSTM layers, so you can stack any LSTM layers you want. For example, you can stack several 2D-LSTM layers and form a deep network, not only in terms of layers, but also in the recurrent dimensions. The intuition is that higher LSTM layers can capture abstract concepts in the sequences, which might help for the task at hand.

Although two continuous events have different timestamp, the event camera can capture the events with extremely high-speed, therefore in practice, the pose of the camera for two continuous event can be considered as the same.

We show by experiments that this formulation is key for good instance segmentation results.

}
\section{Pose Relocalization for Event Camera}\label{Sec:splstm}

\subsection{Problem Formulation}
Inspired by~\cite{Alex2015}~\cite{Alex2016}, we solve the 6DOF pose relocalization task as a regression problem using a deep neural network. Our network is trained to regress a pose vector $ \mathbf{y} = [\mathbf{p}, \mathbf{q}]$ with $\mathbf{p}$ represents the camera position and $\mathbf{q}$ represents the orientation in 3D space. We choose quaternion to represent the orientation since we can easily normalize its four dimensional values to unit length to become a valid quaternion. In practice, the pose vector $\mathbf{y}$ is seven dimensional and is defined relatively to an arbitrary global reference frame. The groundtruth pose labels are obtained through an external camera system~\cite{Mueggler2016_dataset} or structure from motion~\cite{Alex2015}.

\subsection{Stacked Spacial LSTM}
We first briefly describe the Long-Short Term Memory (LSTM) network~\cite{Hochreiter97_LSTM}, then introduce the Stacked Spatial LSTM and the architecture to estimate the 6DOF pose of event cameras. The core of the LSTM is a memory cell which has the gate mechanism to encode the knowledge of previous inputs at every time step. In particular, the LSTM takes an input $\mathbf{x}_t$ at each time step $t$, and computes the hidden state $\mathbf{h}_t$ and the memory cell state $\mathbf{c}_t$ as follows:

\begin{equation}
\begin{aligned} 
{\mathbf{i}_t} &= \sigma ({\mathbf{W}_{xi}}{\mathbf{x}_t} + {\mathbf{W}_{hi}}{\mathbf{h}_{t - 1}} + {\mathbf{b}_i})\\
{\mathbf{f}_t} &= \sigma ({\mathbf{W}_{xf}}{\mathbf{x}_t} + {\mathbf{W}_{hf}}{\mathbf{h}_{t - 1}} + {\mathbf{b}_f})\\
{\mathbf{o}_t} &= \sigma ({\mathbf{W}_{xo}}{\mathbf{x}_t} + {\mathbf{W}_{ho}}{\mathbf{h}_{t - 1}} + {\mathbf{b}_o})\\
{\mathbf{g}_t} &= \phi 	 ({\mathbf{W}_{xg}}{\mathbf{x}_t} + {\mathbf{W}_{hg}}{\mathbf{h}_{t - 1}} + {\mathbf{b}_g})\\
{\mathbf{c}_t} &= {\mathbf{f}_t} \odot {\mathbf{c}_{t - 1}} + {\mathbf{i}_t} \odot {\mathbf{g}_t}\\
{\mathbf{h}_t} &= {\mathbf{o}_t} \odot \phi ({\mathbf{c}_t})
\end{aligned}
\end{equation}
where $\odot$ represents element-wise multiplication; the function $\sigma$ is the sigmoid non-linearity, and $\phi$ is the hyperbolic tangent non-linearity. The weight $\mathbf{W}$ and bias $\mathbf{b}$ are trained parameters. With this gate mechanism, the LSTM network can choose to remember or forget information for long periods of time, while is still robust against vanishing or exploding gradient problems. 

Although the LSTM network is widely used to model temporal sequences, in this work we use the LSTM network to learn spatial dependencies in image feature space. The spatial LSTM has the same architecture as normal LSTM, however, unlike normal LSTM where the input is from the time axis of the data (e.g., a sequence of words in a sentence or a sequence of frames in a video), the input of spatial LSTM is from feature vectors of the image. Recent work showed that the spatial LSTM can further improve the results in many tasks such as music classification~\cite{Keunwoo2016} or image modeling~\cite{Lucas2015}. Stacked Spatial LSTM is simply a stack of several LSTM layers, in which each layer aims at learning the spatial information from image features. The intuition is that higher LSTM layers can capture more abstract concepts in the image feature space, hence improving the results.

\subsection{Pose Relocalization with Stacked Spacial LSTM}
Our pose regression network is composed of two components: a deep CNN and an SP-LSTM network. The CNN network is used to learn deep features from the input event images. After the last layer of the CNN network, we add a dropout layer to avoid overfitting. The output of this CNN network is reshaped and fed to the SP-LSTM module. A fully connected layer is then used to discard the relationships in the output of LSTM. Here, we note that we only want to learn the spatial dependencies in the image features through the input of LSTM, while the relationships in the output of LSTM should be discarded since the components in the pose vector are independent. Finally, a linear regression layer is appended at the end to regress the seven dimensional pose vector. Fig.~\ref{Fig:overview} shows an overview of our approach.

\begin{table*}[!htbp]
\centering\ra{1.3}
\caption{Pose Relocalization Results - Random Split}
\renewcommand\tabcolsep{3.5pt}
\label{tb_result_pose}
\hspace{2ex}

\begin{tabular}{@{}rccccccccc@{}}

\toprule & \multicolumn{2}{c}{PoseNet\cite{Alex2015}}  & \phantom{abc} & \multicolumn{2}{c}{Bayesian PoseNet\cite{Alex2016}}  &  \phantom{abc} & \multicolumn{2}{c}{SP-LSTM (ours)}
\\
\cmidrule{2-3} \cmidrule{5-6} \cmidrule{8-9} & Median Error & Average Error && Median Error & Average Error && Median Error & Average Error
\\


\midrule
\texttt{shapes\_rotation} 				    & $0.109m$, $7.388^\circ$	& $0.137m$, $8.812^\circ$	&& $0.142m$, $9.557^\circ$	& $0.164m$, $11.312^\circ$		&& $0.025m$, $2.256^\circ$	& $0.028m$, $2.946^\circ$ 	\\

\texttt{box\_translation} 				    	& $0.193m$, $6.977^\circ$	& $0.212m$, $8.184^\circ$		&& $0.190m$, $6.636^\circ$	& $0.213m$, $7.995^\circ$			&& $0.036m$, $2.195^\circ$	& $0.042m$, $2.486^\circ$ 	\\
\texttt{shapes\_translation}     				& $0.238m$, $6.001^\circ$	& $0.252m$, $7.519^\circ$		&& $0.264m$, $6.235^\circ$	& $0.269m$, $7.585^\circ$			&& $0.035m$, $2.117^\circ$	& $0.039m$, $2.809^\circ$ 	\\
\texttt{dynamic\_6dof}  				    	& $0.297m$, $9.332^\circ$	& $0.298m$, $11.242^\circ$		&& $0.296m$, $8.963^\circ$	& $0.293m$, $11.069^\circ$			&& $0.031m$, $2.047^\circ$	& $0.036m$, $2.576^\circ$ 	\\
\texttt{hdr\_poster}    				    	& $0.282m$, $8.513^\circ$	& $0.296m$, $10.919^\circ$		&& $0.290m$, $8.710^\circ$	& $0.308m$, $11.293^\circ$			&& $0.051m$, $3.354^\circ$	& $0.060m$, $4.220^\circ$ 	\\
\texttt{poster\_translation}    		    	& $0.266m$, $6.516^\circ$	& $0.282m$, $8.066^\circ$		&& $0.264m$, $5.459^\circ$	& $0.274m$, $7.232^\circ$			&& $0.036m$, $2.074^\circ$	& $0.041m$, $2.564^\circ$ 	\\
\cline{1-9}
\textbf{Average}							    & $0.231m$, $7.455^\circ$	& $0.246m$, $9.124^\circ$		&& $0.241m$, $7.593^\circ$	& $0.254m$, $9.414^\circ$		&& $0.036m$, $2.341^\circ$	& $0.041m$, $2.934^\circ$  \\
\bottomrule
\end{tabular}
\end{table*}

In practice, we choose the VGG16~\cite{SimonyanZ14} network as our CNN. We first discard its last softmax layer and add a dropout layer with the rate of $0.5$ to avoid overfitting. The event image features are stored in the last fully connected layer in a $4096$ dimensional vector. We reshape this vector to $64 \times 64$ in order to feed to the LSTM module with $256$ hidden units. Here, we can consider that the inputs of LSTM are from $64$ ``feature sentences", each has $64$ ``words", and the spatial dependencies are learned from these sentences. We then add another LSTM network to create an SP-LSTM with $2$ layers. The output of SP-LSTM module is fed to a fully connected layer with $512$ neurons, following by another fully connected layer with $7$ neurons to regress the pose vector. We choose the SP-LSTM network with $2$ layers since it is a good balance between accuracy and training time.

\subsection{Training}
To train the network end-to-end, we use the following objective loss function:

\begin{equation}
\mathcal{L}(I) = {\left\| {\mathbf {\hat p} - \mathbf{p}} \right\|_2} + {\left\| {\mathbf{\hat q} - \mathbf{q}} \right\|_2}
\end{equation}
where $\mathbf{\hat p}$ and $\mathbf{\hat q}$ are the predicted position and orientation from the network. In~\cite{Kendall2017}, the authors proposed to use a geometry loss function to encode the spatial dependencies from the input. However, this approach required a careful initialization and needed a list of 3D points to measure the projection error of the estimated pose, which is not available in the groundtruth of the dataset we use in our experiment.

For simplicity, we choose to normalize the quaternion to unit length during testing phrase, and use Euclidean distance to measure the difference between two quaternions as in~\cite{Alex2015}. Theoretically, this distance should be measured in spherical space, however, in practice the deep network outputs the predicted quaternion $\mathbf{\hat q}$ close enough to the groundtruth quaternion $\mathbf{q}$, making the difference between the spherical and Euclidean distance insignificant. We train the network for $1400$ epochs using stochastic gradient descent with $0.9$ momentum and $1e-6$ weight decay. The learning rate is empirically set to $1e-5$ and kept unchanging during the training. It takes approximately $2$ days to train the network from scratch on a Tesla P100 GPU.

\junk{
 Although in theory, this distance should be measured in spherical space, in practice the deep network outputs the predicted quaternion $\mathbf{\hat q}$ close enough to the groundtruth quaternion $\mathbf{q}$. This makes the difference between the spherical and Euclidean distance becomes insignificant.

The intuition is that the LSTM network would learn the spatial dependencies from a $64$ length vector easier than from a $4096$ length vector.
 
The whole network is trained end-to-end through back propagation.

In practice, we can use any state-of-the-art deep architecture designed for image classification task as our CNN.

We train a first sequence from scratch and then use its weights to initialize the weights of the other sequences. It takes approximately $3$ days to train the first network from scratch on a Titan X GPU, while using pretrained weights reduces the training time for other sequences to approximately $1$ day.

A dropout layer is used after the CNN network to avoid overfitting.

In practice, we use the state-of-the-art deep CNN to extract these feature vectors from the input images.

$e_i = <t_i, (x_i, y_i), \rho_i>$
It also has higher high dynamic range in comparison with the frame-based cameras ($130$ dB vs. $60$ dB).

Since the network need to learn the translation and rotation of the pose vector at the same time, while these two components are measured in different units, it is not trivial to design an effective loss function. To train the network, we used the following objective loss function:

\begin{equation}
\mathcal{L}(I) = {\left\| {\mathbf {\hat x} - \mathbf{x}} \right\|_2} + \beta {\left\| {\mathbf{\hat q} - \frac{\mathbf{q}}{{\left\| \mathbf{q} \right\|}}} \right\|_2}
\end{equation}

where $\beta$ is a parameter to keep the loss from the position and orientation to be approximately equal. For simplicity, the orientation loss is calculated on Euclidean space instead of the unit sphere in quaternion space since in practice the distinction between spherical distance and Euclidean distance is insignificant~\cite{•}. In~\cite{•}, the authors introduced a new loss function based on the reprojection error. However, this approach needed a list of 3D points to measure the projection error with the estimated pose $\mathbf{\hat p}$, which is not available in the groundtruth of our dataset.

(three dimensions represent position $\mathbf{x}$ and four dimensions represent orientation $\mathbf{q}$),

In practice, the LSTM network is straightforward to train end-to-end and can handle inputs with different lengths using the padding techniques.

To demonstrate the generality of our approach, we instantiate Mask R-CNN with multiple architectures. For

This allows each layer in the mask branch to maintain the explicit m × m object spatial lay- out without collapsing it into a vector representation that lacks spatial dimensions. Unlike

Even though Long Short-Term Memory (LTSM) units
have been typically applied to temporal sequences, recent works [6, 32, 53, 54] have used the memory capabilities of LSTMs in image space. In our case, we treat the output 2048 feature vector as our sequence. We propose to insert four LSTM units after the FC, which have the function of reducing the dimensionality of the feature vector in a struc- tured way. The memory units filter out the most useful fea- ture correlations for the task of pose estimation

Stack of LSTM, on other hands, is simply a stack of several LSTM layers, so you can stack any LSTM layers you want. For example, you can stack several 2D-LSTM layers and form a deep network, not only in terms of layers, but also in the recurrent dimensions. The intuition is that higher LSTM layers can capture abstract concepts in the sequences, which might help for the task at hand.

Although two continuous events have different timestamp, the event camera can capture the events with extremely high-speed, therefore in practice, the pose of the camera for two continuous event can be considered as the same.

We show by experiments that this formulation is key for good instance segmentation results.

}
\section{EXPERIMENTS} \label{Sec:exp}
\subsection{Dataset}

We use the event camera dataset that was recently introduced in~\cite{Mueggler2016_dataset} for our experiment. This dataset included a collection of scenes captured by a DAVIS camera in indoor and outdoor environments. The indoor scenes of this dataset have the groundtruth camera poses from a motion-capture system with sub-millimeter precision at $200$ Hz. We use the timestamp of the motion-capture system to create event images. All the events with the timestamp between $t$ and $t+1$ of the motion-capture system are grouped as one event image. Without the loss of generality, we consider the groundtruth pose of this event image is the camera pose that was taken by the motion-capture system at time $t+1$. This assumption technically limits the speed of the event camera to the speed of the motion-capture system (i.e. $200$ Hz), however it allows us to use the groundtruth poses with sub-millimeter precision from the motion-capture system.

\textbf{Random Split} As the standard practice in the pose relocalization task~\cite{Alex2015}, we \textit{randomly} select $70\%$ of the event images for training and the remaining $30\%$ for testing. We use $6$ sequences (\texttt{shapes\_rotation}, \texttt{box\_translation}, \texttt{shapes\_translation}, \texttt{dynamic\_6dof}, \texttt{hdr\_poster}, \texttt{poster\_translation}) for this experiment. These sequences are selected to cover different camera motions and scene properties.

\textbf{Novel Split} To demonstrate the generalization ability of our SP-LSTM network, we also conduct the experiment using the novel split. In particular, from the original event images sequence, we select \textit{the first} $70\%$ of the event images for training, then \textit{the rest} $30\%$ for testing. In this way, we have two independent sequences on the same scene (i.e., the training sequence is selected from timestamp $t_0$ to $t_{70}$, and the testing sequence is from timestamp $t_{71}$ to $t_{100}$). We use three sequences from the \texttt{shapes} scene (\texttt{shapes\_rotation}, \texttt{shapes\_translation}, \texttt{shapes\_6dof}) in this novel split experiment to compare the results when different camera motions are used.

We note that in both the random split and novel split strategies, after having the training/testing set, our SP-LSTM network selects the event image randomly for training/testing, and no sequential information between event images is needed. Moreover, unlike the methods in~~\cite{zihao2017event}~\cite{Rebecq2017} that need the inertial measurement data, our SP-LSTM \textit{only} uses the event images as the input.

\begin{table*}[!htbp]
\centering\ra{1.3}
\caption{Pose Relocalization Results - Novel Split}
\renewcommand\tabcolsep{2.5pt}
\label{tb_novel_split_result}
\hspace{2ex}

\begin{tabular}{@{}rccccccccc@{}}
\toprule & \multicolumn{2}{c}{PoseNet\cite{Alex2015}}  & \phantom{abc} & \multicolumn{2}{c}{Bayesian PoseNet\cite{Alex2016}}  &  \phantom{abc} & \multicolumn{2}{c}{SP-LSTM (ours)}
\\
\cmidrule{2-3} \cmidrule{5-6} \cmidrule{8-9} & Median Error & Average Error && Median Error & Average Error && Median Error & Average Error
\\ 

\midrule
\texttt{shapes\_rotation} 				    & $0.201m$, $12.499^\circ$	& $0.214m$, $13.993^\circ$	&& $0.164m$, $12.188^\circ$	& $0.191m$, $14.213^\circ$		&& $0.045m$, $5.017^\circ$	& $0.049m$, $11.414^\circ$ 	\\

\texttt{shapes\_translation} 				& $0.198m$, $6.969^\circ$	& $0.222m$, $8.866^\circ$		&& $0.213m$, $7.441^\circ$	& $0.228m$, $10.142^\circ$		&& $0.072m$, $4.496^\circ$	& $0.081m$, $5.336^\circ$ 	\\

\texttt{shapes\_6dof}     				& $0.320m$, $13.733^\circ$	& $0.330m$, $18.801^\circ$		&& $0.326m$, $13.296^\circ$	& $0.329m$, $18.594^\circ$		&& $0.078m$, $5.524^\circ$	& $0.095m$, $9.532^\circ$ 	\\

\cline{1-9}
\textbf{Average}					    & $0.240m$, $11.067^\circ$	& $0.255m$, $13.887^\circ$		&& $0.234m$, $10.975^\circ$	& $0.249m$, $14.316^\circ$		&& $0.065m$, $5.012^\circ$	& $0.075m$, $8.761^\circ$  \\
\bottomrule
\end{tabular}
\end{table*}

\subsection{Baseline}
We compare our experimental results with two recent state-of-the-art methods in computer vision: PoseNet~\cite{Alex2015} and Bayesian PoseNet~\cite{Alex2016}. We note that both our SP-LSTM, PoseNet and Bayesian PoseNet use only the event images as the input and no further information such as 3D map of the environment or inertial measurements is needed.

For each sequence, we report the median and average error of the estimated poses in position and orientation separately. The predicted position is compared with the groundtruth using the Euclidean distance, while the predicted orientation is normalized to unit length before comparing with the groundtruth. The median and average error are measured in $m$ and $deg$ for the position and orientation, respectively.

\subsection{Random Split Results}

Table~\ref{tb_result_pose} summarizes the median and average error on $6$ sequences using the random split strategy. From this table, we notice that the pose relocalization results are significantly improved using our SP-LSTM network in comparison with the baselines that used only CNN~\cite{Alex2015}~\cite{Alex2016}. Our SP-LSTM achieves the lowest mean and average error in all sequences. In particular, SP-LSTM achieves $0.036m$, $2.341^\circ$ in median error on average of all sequences, while PoseNet and Bayesian PoseNet results are $0.231m$, $7.455^\circ$ and $0.241m$, $7.593^\circ$, respectively. Overall, this improvement is around $6$ times in position error and $3$ times in orientation error. This demonstrates that the spatial dependencies play an important role in the camera pose relocalization process and our SP-LSTM successfully learns these dependencies, hence significantly improves the results. We also notice that PoseNet performs slightly better than Bayesian PoseNet, and the uncertainty estimation in Bayesian PoseNet cannot improve the pose relocalization results for event data.

From Table~\ref{tb_result_pose}, we notice that the pose relocalization results also depend on the properties of the scene in each sequence. Due to the design mechanism of the event-based camera, the events are mainly captured around the contours of the scene. In cluttered scenes, these contours are ambiguous due to non-meaningful texture edge information. Therefore, the event images created from events in these scenes are very noisy. As the results, we have observed that for sequences in cluttered or dense scenes (e.g. \texttt{hdr\_poster}), the pose relocalization error is higher than sequences from the clear scenes (e.g. \texttt{shapes\_rotation}, \texttt{shapes\_translation}). We also notice that dynamic objects (e.g. as in \texttt{dynamic\_6dof} scene) also affect the pose relocalization results. While PoseNet and Bayesian Posenet are unable to handle the dynamic objects and have high position and orientation errors, our SP-LSTM gives reasonable results in this sequence. It demonstrates that by effectively learning the spatial dependencies with SP-LSTM, the results in such difficult cases can be improved.

\begin{figure}[ht]
  \centering
    \subfigure[]{\label{fig:a}\includegraphics[width=0.48\linewidth, height=0.4\linewidth]{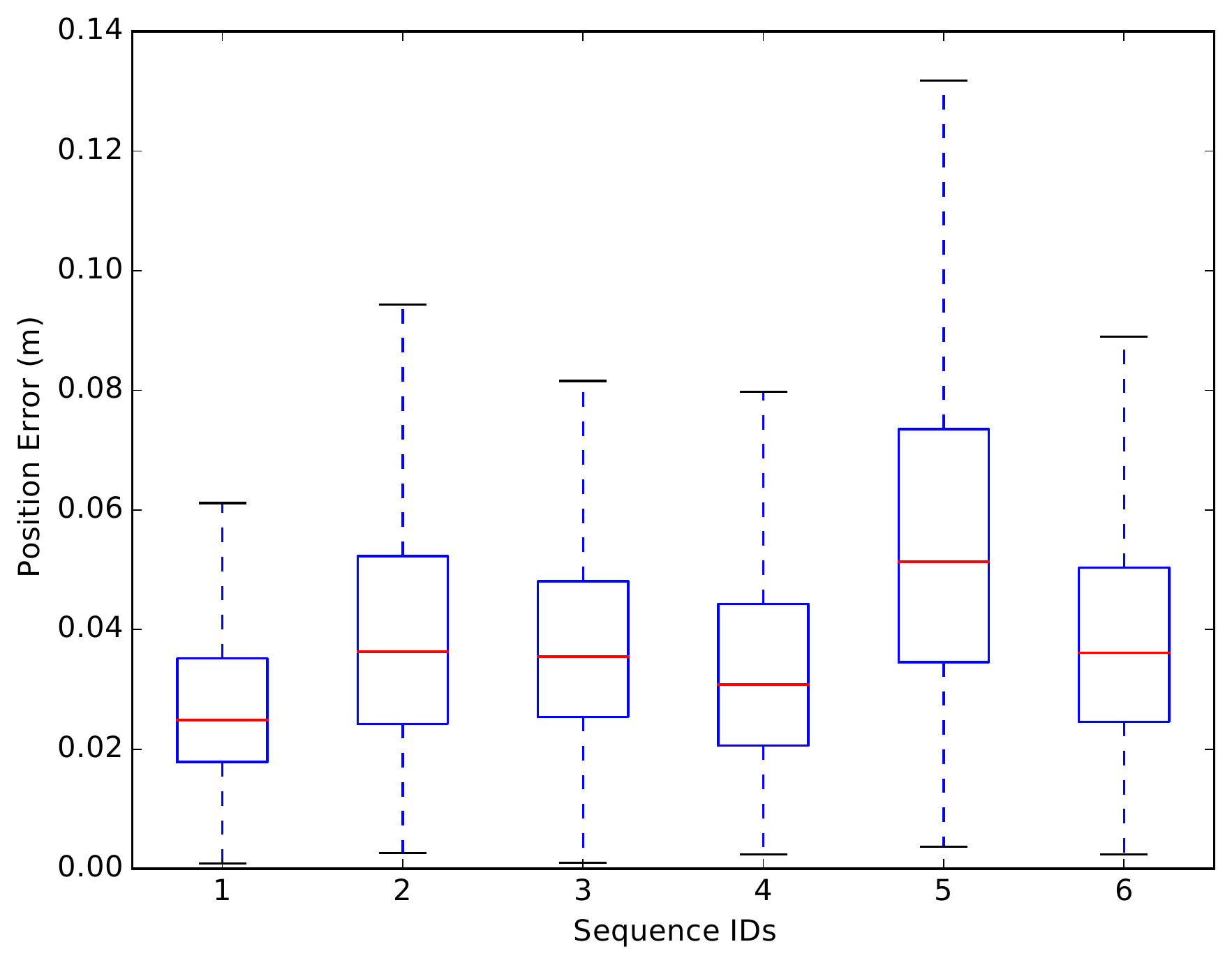}}
    \hspace{0ex}
    \subfigure[]{\label{fig:b}\includegraphics[width=0.48\linewidth, height=0.4\linewidth]{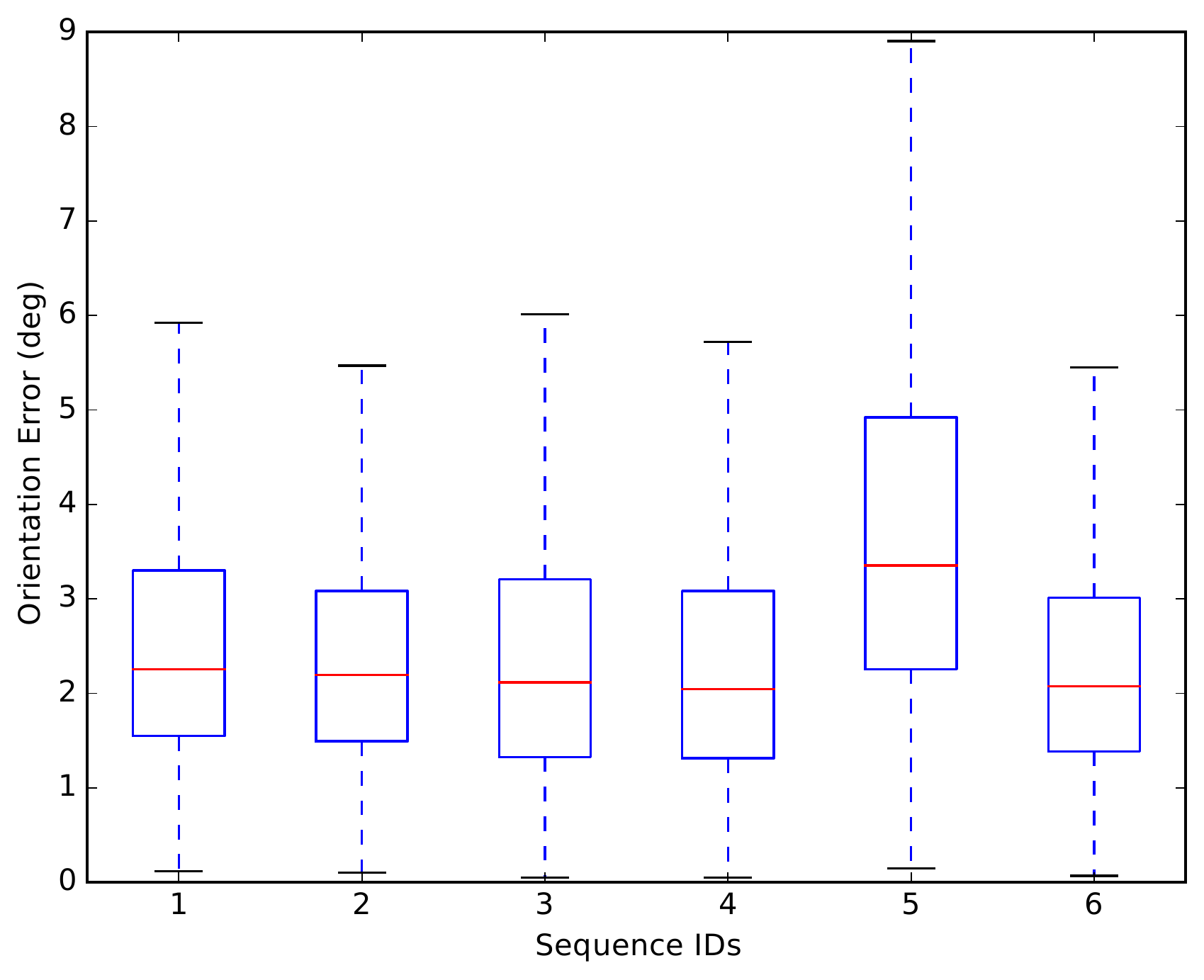}}
     
 \vspace{2.0ex}
 \caption{Error distribution of the pose relocalization results of our SP-LSTM network using random split. \textbf{(a)} Position error distribution. \textbf{(b)} Orientation error distribution.  Sequences IDs are: 1-\texttt{shapes\_rotation}, 2-\texttt{box\_translation}, 3-\texttt{shapes\_translation}, 4-\texttt{dynamic\_6dof}, 5-\texttt{hdr\_poster}, 6-\texttt{poster\_translation}.}
 \label{Fig:result_error_distribution}
  \vspace{0.5ex}
\end{figure}

\textbf{Error Distribution} Fig.~\ref{Fig:result_error_distribution} shows the position and orientation error distributions of our SP-LSTM network. Each box plot represents the error for one sequence. We recall that the top and bottom of a box are the first and third quartiles that indicate the interquartile range (IQR). The band inside the box is the median. We notice that the IQR of position error of all sequences (except the \texttt{hdr\_poster}) is around $0.02m$ to $0.05m$, while the maximum error is around $0.09m$. The IQR of orientation error is in the range $1.5^\circ$ to $3.5^\circ$, and the maximum orientation error is only $6^\circ$. In all $6$ sequences in our experiment, the \texttt{hdr\_poster} gives the worst results. This is explainable since this scene is a dense scene, hence the event images have uncleared structure and very noisy. Therefore, it is more difficult for the network to learn and predict the camera pose from these images.

\subsection{Novel Split Results}

Table~\ref{tb_novel_split_result} summarizes the median and average error on $3$ sequences using the novel split strategy. This table clearly shows that our SP-LSTM results outperform both PoseNet and Bayesian PoseNet by a substantial margin. Our SP-LSTM achieves the lowest median and average error in both $3$ sequences in this experiment, while the errors of PoseNet and Bayesian PoseNet remain high. In particular, the median error of our SP-LSTM is only $0.065m$ and $5.012^\circ$ in average, compared to $0.240m$, $11.067^\circ$ and $0.234m$, $10.975^\circ$ from PoseNet and Bayesian PoseNet errors, respectively. These results confirm that by learning the spatial relationship in the image feature space, the pose relocalization results can be significantly improved. Table~\ref{tb_novel_split_result} also shows that the domination motion of the sequence also affects the results, for example, the translation error in the \texttt{shapes\_translation} sequence is higher than \texttt{shapes\_rotation}, and vice versa for the orientation error.

Compared to the pose relocalization errors using the random split (Table~\ref{tb_result_pose}), the relocalization errors using the novel split are generally higher. This is explainable since the testing set from the novel split is much more challenging. We recall that in the novel split, the testing set is selected from the last $30\%$ of the event images. This means we do not have the ``neighborhood" relationship between the training and testing images. In the random split strategy, the testing images can be very close to the training images since we select the images randomly from the whole sequence for training/testing. This does not happen in the novel split strategy since the training and testing set are two separated sequences. Despite this challenging setup, our SP-LSTM still is able to regress the camera pose and achieves reasonable results. This shows that the network successfully encodes the geometry of the scene during training, hence generalizes well during the testing phase.

To conclude, the extensive experimental results from both the random split and novel split setup show that our SP-LSTM network successfully relocalizes the event camera pose using only the event image. The key reason that leads to the improvement is the use of stacked spatial LSTM to learn the spatial relationship in the image feature space. The experiments using the novel split setup also confirm that our SP-LSTM successfully encodes the geometry of the scene during the training and generalizes well during the testing. Furthermore, our SP-LSTM also has very fast inference time and requires only the event image as the input to relocalize the camera pose.

\textbf{Reproducibility} We implement the proposed method using Tensorflow framework~\cite{TensorFlow2015_short}. The testing time for each new event image using our implementation is around $5ms$ on a Tesla P100 GPU, which is comparable to the real-time performance of PoseNet, while the Bayesian PoseNet takes longer time (approximately $240ms$) due to the uncertainty analysis process. To encourage further research, we will release our source code and trained models that allow reproducing the results in this paper.

\subsection{Robustness to Number of Events }
\begin{figure}[ht]
  \centering
    \subfigure[]{\label{fig:a}\includegraphics[width=0.48\linewidth, height=0.4\linewidth]{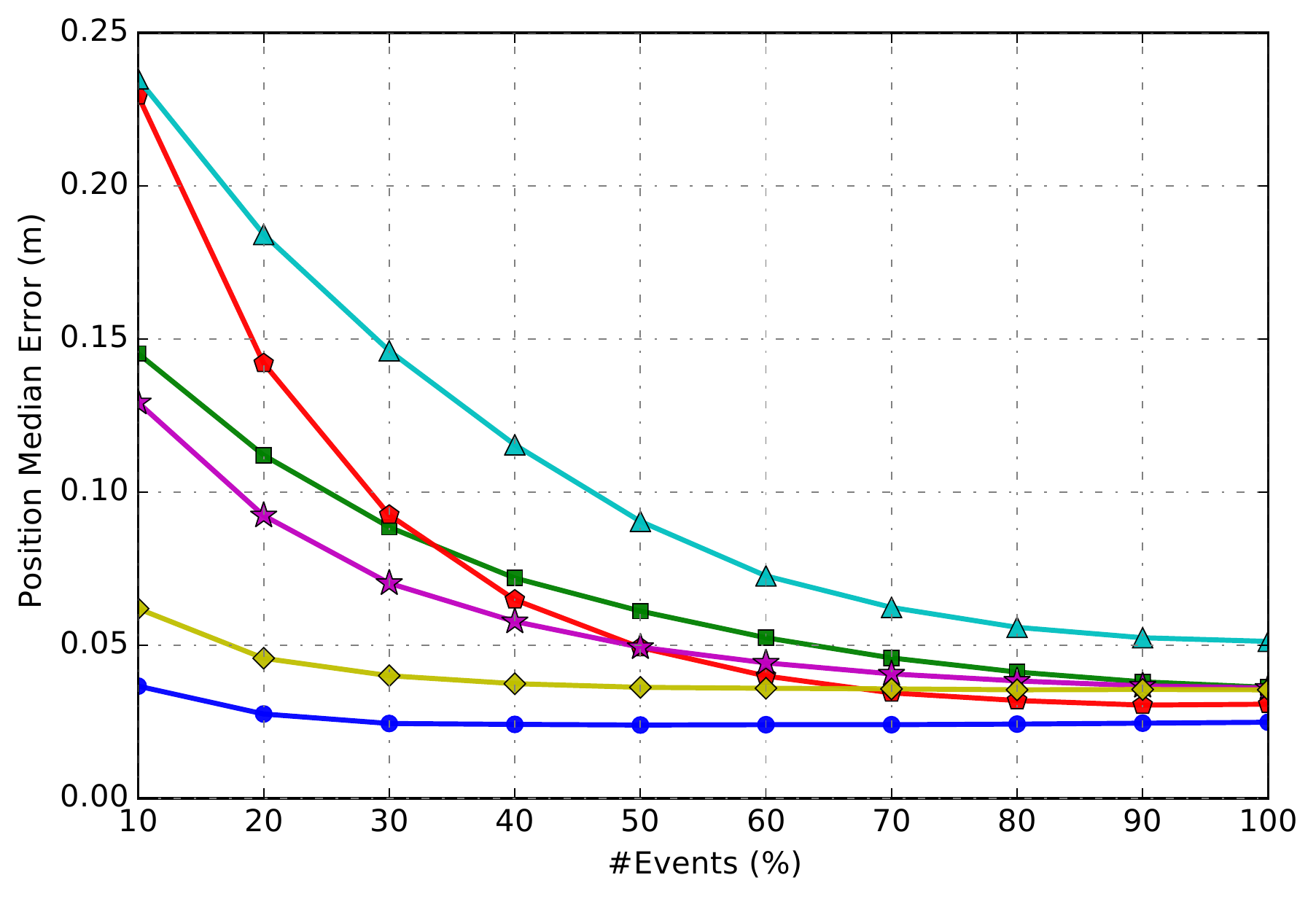}}
    \hspace{0ex}
    \subfigure[]{\label{fig:b}\includegraphics[width=0.48\linewidth, height=0.4\linewidth]{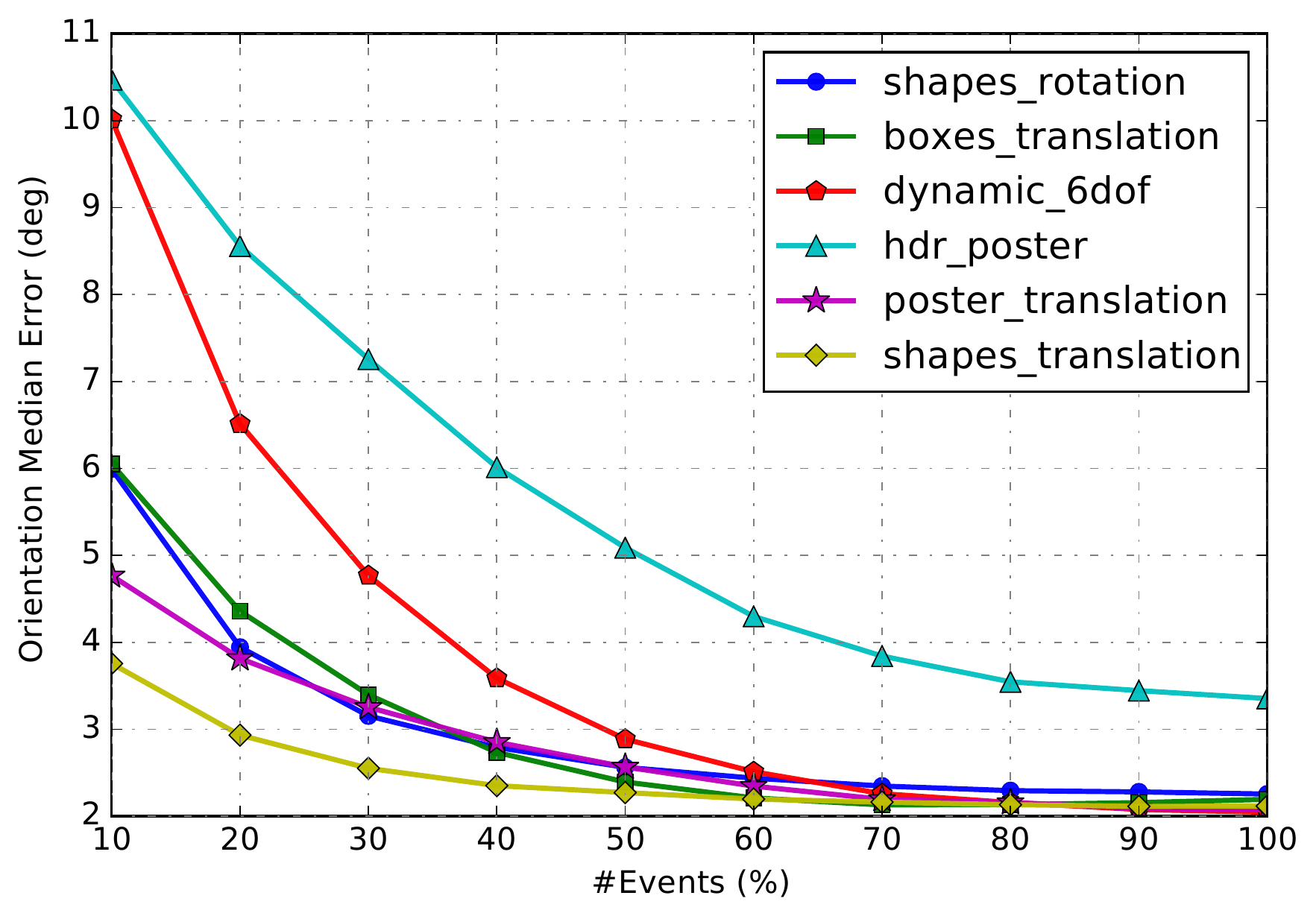}}
     
 \vspace{2.0ex}
 \caption{Robustness to number of events. \textbf{(a)} Position median error. \textbf{(b)} Orientation median error. The position and orientation errors of our SP-LSTM network do not significantly drop when we use more than $60\%$ of all events to create the event images.}
 \label{Fig:result_num_event}
\end{figure}

In this work, we assume that $n$ events occurring between two timestamps of the external camera system will have the same camera pose. Although this assumption is necessary to use the groundtruth poses to train the network, it limits the speed of the event-based camera to the sampling rate of the external camera system. To analyze the effect of number of events to the pose relocalization results, we perform the following study: During the testing phase using the random split strategy, instead of using all $100\%$ events from two continuous timestamps, we gradually use only $10\%$, $20\%$, ..., $90\%$ of these events to create the event images (the events are chosen in order from the current timestamp to the previous timestamp). Fig.~\ref{Fig:result_num_event} shows the position and orientation errors of our SP-LSTM network in this experiment. From the figure, we notice that both the position and orientation errors of all sequences become consistent when we use around $60\%$ number of events. When we use more events to create the event images, the errors are slightly dropped but not significantly. This suggests that the SP-LSTM network still performs well when we use fewer events. We also notice that our current method to create the event image from the events is fairly simple since some of the events may be overwritten when they occur at the same coordinates but have different polarity values with the previous events. Despite this, our SP-LSTM network still successfully relocalizes the camera pose from the event images.

\junk{

 , where we can qualitatively see that the estimated trajectory is very similar to the ground truth.

Both PoseNet and Bayesian PoseNet use the GoogleNet~\cite{Szegedy16_Inception} as the main CNN backbone to regress the camera pose. We generally follow the training process described in the associated papers and use the source code provided by the authors for a fair comparison.

In addition to the asynchronous event data, this dataset also provides the global-shutter intensity images and inertial measurements.



upon acceptance

To the best of our knowledge, our work is the first success when using deep learning to estimate the pose of the event cameras.

This will also help us to have a fair comparison between the method that used learning approach (such as ours) and the others that used geometry approach~\cite{•} for this problem in the future.

 This is expected since the indoor scenes only cover a small area and the dynamic objects become more important in the scene. We expect that this problem will not occur for outdoor scenes with large area since the network can learn to

.  It demonstrates that by using the LSTM network to learn spatial dependencies in feature space produced by CNN, the pose estimation error can be significantly reduced in both position and orientation aspect.

We first train the \texttt{shapes\_rotation} sequence from scratch then use its weight to train the other sequences. 
 
During the testing phrase, we also show that our method is robust again the number of events when forming an event images.
  
Other groundtruth generation methods such as using structure from motion have higher precision errors~\cite{Alex2015}.
}
\section{Conclusions and Future Work}\label{Sec:con}
In this paper, we introduce a new method to relocalize the 6DOF pose of the event camera with a deep network. We first create the event images from the event stream. A deep convolutional neuron network is then used to learn features from the event image. These features are reshaped and fed to a Stacked Spatial LSTM network. We have demonstrated that by using the Stacked Spatial LSTM network to learn spatial dependencies in the feature space, the pose relocalization results can be significantly improved. The experimental results show that our network generalizes well under challenging testing strategies and also gives reasonable results when fewer events are used to create event images. Furthermore, our method has fast inference time and needs only the event image to relocalize the camera pose.

Currently, we employ a fairly simple method to create the event image from a list of events. Our forming method does not check if the event at the local pixel has occurred or not. Since the input of the deep network is the event images, better forming method can improve the pose relocalization results, especially on the cluttered scenes since the data from event cameras are very disorder. Although our network achieves $5ms$ inference time, which can be considered as real-time performance as in PoseNet, it still may not fast enough for high-speed robotic applications using event cameras. Therefore, another interesting problem is to study the compact network architecture that can achieve competitive pose relocalization results while having fewer layers and parameters. This would improve the speed of the network and allow it to be used in more realistic scenarios.

\vspace{2ex}
\section*{Acknowledgment}
\addcontentsline{toc}{section}{Acknowledgment}
Anh Nguyen, Darwin G. Caldwell and Nikos G. Tsagarakis are supported by the European Union Seventh Framework Programme (FP7-ICT-2013-10) under grant agreement no 611832 (WALK-MAN). Thanh-Toan Do is supported by the Australian Research Council through the Australian Centre for Robotic Vision (CE140100016).

\junk
{

works directly on event stream and 

Currently, our approach needs two separate networks to detect object affordances. This architecture can't be trained end-to-end as a single network. In future work, we aim to overcome this limitation by developing a new architecture that can detect the object identity and its affordances simultaneously. Another interesting problem is to extend our robotics experiments with more complicated scenarios.
}

\bibliographystyle{class/IEEEtran}
\bibliography{class/IEEEabrv,class/reference}
   
\end{document}